\documentclass[letterpaper, 10 pt, conference]{ieeeconf}  

\IEEEoverridecommandlockouts                              

\usepackage{amsmath,amsfonts}
\usepackage{algorithm}
\usepackage{algpseudocode}
\usepackage{caption}
\usepackage{array}
\usepackage[caption=false,font=normalsize,labelfont=sf,textfont=sf]{subfig}
\usepackage{textcomp}
\usepackage{stfloats}
\usepackage{url}
\usepackage{verbatim}
\usepackage{graphicx}
\usepackage{multirow}
\usepackage{newtxtext,newtxmath}
\usepackage{booktabs}
\usepackage{cite}
\usepackage{hyperref}
\usepackage[color=green!20, textsize=footnotesize]{todonotes} 

\definecolor{commentcolor}{HTML}{00e268}%

\makeatletter
\renewcommand{\ALG@beginalgorithmic}{\small}
\algrenewcommand\algorithmiccomment[1]{\hfill\textcolor{blue}{$\triangleright$ #1}}
\makeatother

\author{ Yubin Ke$^{1, 2*}$, Jiayi Chen$^{1, 2*}$,Hang Lv$^{1, 2*}$, Xiao Zhou$^{1, 2}$ and He Wang$^{1, 2, 3\dagger}$
\thanks{$^{1}$Peking University. $^{2}$Galbot. $^{3}$Beijing Academy of Artificial Intelligence.}%
\thanks{$^*$Equal contribution. $^\dagger$Corresponding author: \href{mailto:hewang@pku.edu.cn}{hewang@pku.edu.cn}.}
}
\title{\LARGE \bf TacDexGrasp: Compliant and Robust \\ Dexterous  Grasping with Tactile Feedback}

\begin{document}

\maketitle
\thispagestyle{empty}
\pagestyle{empty}

\begin{abstract}
Multi-fingered hands offer great potential for compliant and robust grasping of unknown objects, 
yet their high-dimensional force control presents a significant challenge. 
This work addresses two key problems: (1) distributing forces across multiple contacts to counteract an object's weight, 
and (2) preventing rotational slip caused by gravitational torque when a grasp is distant from the object's center of mass. 
We address these challenges via tactile feedback and a Second-Order Cone Programming (SOCP)-based controller, 
without explicit torque modeling or slip detection. 
Our key insights are (1) rotational slip inevitably induces translational slip at some contact points for a multi-fingered grasp, 
and (2) the ratio of tangential to normal force at each contact is an effective early stability indicator. 
By actively constraining this ratio for each finger below the estimated friction coefficient, 
our controller maintains grasp stability against both translational and rotational slip. 
Real-world experiments on 12 diverse objects demonstrate the robustness and compliance of our approach.
\end{abstract}

\section{INTRODUCTION}

Applying appropriate forces is essential for both stability and safety in robotic grasping. 
Excessive force may damage delicate objects, while insufficient force can lead to slippage or grasp failure. 
This trade-off is particularly critical in real-world scenarios such as service robotics and human-robot collaboration, 
where the robot must securely grasp objects of unknown physical properties without causing damage. Achieving such compliant yet robust grasping requires not only reliable perception of contact interactions—often provided by tactile sensors—but also precise real-time force control.

Prior work on compliant grasping~\cite{hang2016hierarchical,romano2011slip,ajoudani2016slip,muthusamy2020slip,huang2024newrelated,cui2023slip,chen2025slip,James2021slip,pasluosta2009slippage,zhang2016initial,ford2025shear,wettels2009force,zhang2014force,li2025RAL,sui2023muest}
has largely focused on parallel grippers or highly underactuated multi-fingered hands.
These designs simplify control by reducing the number of degrees of freedom (DoF) but sacrifice dexterity and adaptability.
For example, parallel grippers are particularly prone to rotational slip caused by gravitational torque when the grasp is offset from the object’s center of mass.

In contrast, fully actuated dexterous hands—with their human-like morphology—hold greater potential for versatile and robust grasping. 
However, this potential remains underexplored due to the challenges introduced by their high DoF and complex contact interactions.
Specifically, two key problems must be addressed: (1) how to optimally distribute contact forces across multiple contact points, and (2) how to effectively prevent rotational slip.

\begin{figure}[ht]
  \centering
  \includegraphics[width=0.9\linewidth]{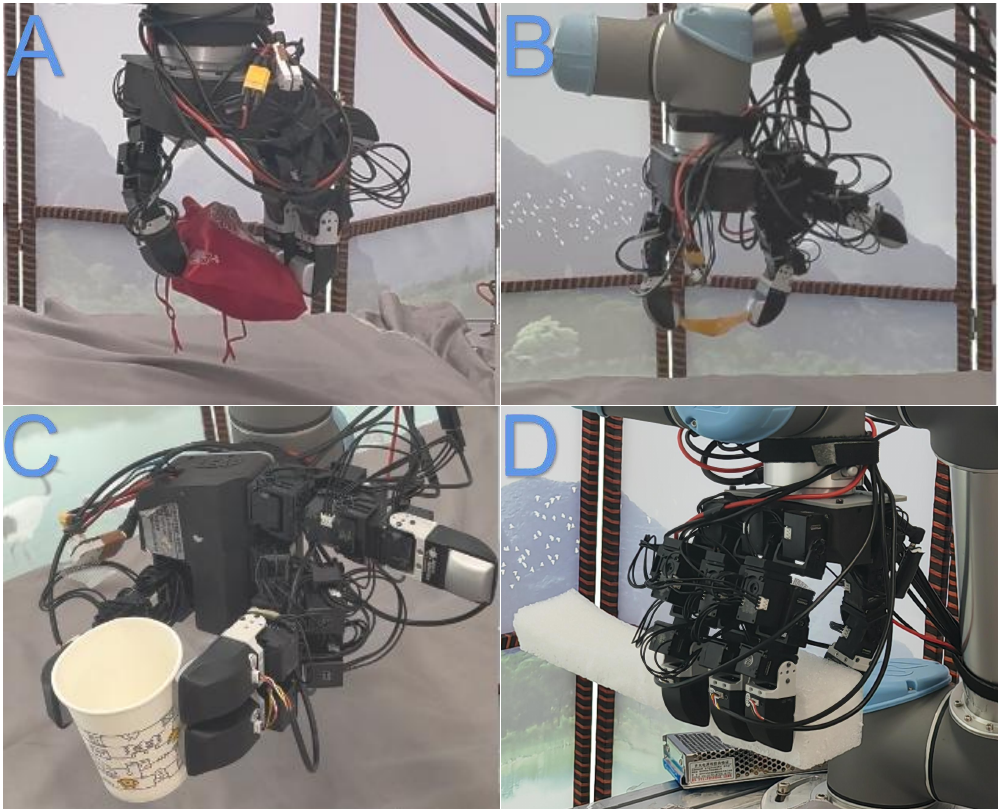} 
  \caption{\textbf{TacDexGrasp.} (A-C) Our system enables stable and safe grasping for multi-fingered hands on objects with diverse, 
  unknown mass distributions, friction coefficients, and deformation materials. 
  (D) Our SOCP-based controller can compensate for gravitational torque to prevent the rotational slip without explicit torque modeling.}
  \label{fig:teaser}
\end{figure}

In this work, we tackle these challenges using tactile feedback and a Second-Order Cone Programming (SOCP)-based force controller, 
without relying on explicit torque modeling or slip detection. 
Our approach is motivated by two key insights:

First, rotational slip of an object induces translational slip at certain contacts in a multi-fingered grasp, as shown in our Method section.
This relationship holds strictly for rigid objects when the contact points are not collinear, and, under mild assumptions, 
also extends to deformable objects in most practical cases.

Second, the ratio of tangential to normal force at a contact provides an early and reliable indicator of grasp stability.
By constraining this ratio below the estimated friction coefficient, slip can be avoided before it occurs.
This constraint can be naturally integrated into a SOCP formulation, which we solve efficiently in around 10 ms to compute the optimal force distribution across contacts. 
The resulting target forces are then tracked using a PID controller on the robot hand.

Our method is validated in real-world experiments on 12 objects with diverse, unknown mass distributions, friction coefficients, and deformation materials. 
Our approach achieves an $83\%$ grasp success rate with $38\%$ lower contact forces without damaging delicate objects, outperforming prior baselines, especially on deformable objects and elongated objects. 
In addition to improved average performance, our controller also demonstrates robustness against external disturbances such as mass variations, external accelerative disturbances, and irregular arm motions.

To summarize, our contributions are:
\begin{itemize}
\item A novel tactile-based SOCP formulation for multi-fingered grasping that jointly prevents translational and rotational slip, without explicit slip detection or torque modeling.
\item A SOCP-based adaptive control system for compliant and robust grasping of unknown objects.
\item Real-world experiments on 12 diverse objects demonstrating the effectiveness and robustness of our approach.
\end{itemize}

\section{RELATED WORK}

\subsection{Tactile Sensing in Robotics}

Tactile sensing is fundamental for enabling robotic systems to interact safely and adaptably with unstructured environments. While earlier grasping methods relied heavily on binary contact sensors or purely normal force measurements, modern dexterous manipulation increasingly demands high-fidelity, multi-axis force feedback to detect shear forces and prevent slip~\cite{yuan2017gelsight,ward2018tactip}. Various transduction principles—such as piezoresistive, capacitive, and optical—have been developed to capture these complex contact dynamics. Regardless of the underlying mechanism, the critical requirement for real-time force control is the reliable estimation of contact wrenches. In this work, our approach is agnostic to the specific sensor type, provided it yields accurate resultant force vectors. For our hardware implementation, we utilize commercially available Tac3D sensors~\cite{zhang2022tac3d} due to their stability, high spatial resolution, and the real-time 3D force feedback necessary for our closed-loop controller.

\subsection{Convex Optimization in Robotic Grasping}

Mathematical optimization has long served as a foundational framework for resolving multi-fingered grasp forces~\cite{martin1996dextrous, han2000grasp}. While Second-Order Cone Programming (SOCP) represents the mathematical standard for accurately modeling the non-linear Coulomb friction cone, many recent works~\cite{chen2024bodex,wu2022learning,chen2025dexonomy} approximate these quadratic cones as polyhedrons to improve computational efficiency. This approximation transforms the exact SOCP into a computationally lighter Quadratic Programming (QP) problem, but sacrifices the exactness of the friction constraints. Furthermore, these optimization frameworks are predominantly employed as objective functions for offline grasp synthesis, rather than for closed-loop execution. However, the advent of highly efficient modern solvers, such as Clarabel~\cite{Clarabel_2024}, now enables exact SOCP formulations to be solved in real time. Leveraging this capability, our work integrates high-frequency SOCP with dynamic tactile feedback, successfully transitioning optimal grasp force allocation into a closed-loop control system to achieve compliant and robust grasping.

\subsection{Tactile-Based Grasping}
Existing tactile grasping algorithms primarily rely on either slip-based modulation or reactive force control. Earlier slip-based studies~\cite{romano2011slip,ajoudani2016slip,muthusamy2020slip,huang2024newrelated} utilized heuristic rules for slip detection, whereas recent learning-based approaches~\cite{cui2023slip,chen2025slip,James2021slip,pasluosta2009slippage,zhang2016initial} utilize tactile data to improve detection accuracy and generalization. 
Parallelly, force-based methods~\cite{ford2025shear,wettels2009force,zhang2014force} monitor variations in normal and shear forces to adjust the grasp reactively. 
Recent learning-based tactile controllers~\cite{li2025RAL,sui2023muest,hogan2018tactile} further incorporate tactile priors to accelerate adaptation. 
However, these approaches are predominantly developed for low-DoF grippers or underactuated hands, often failing to fully exploit the control redundancy and dexterity offered by fully actuated multi-fingered hands.

Tactile-based grasping for fully actuated dexterous hands remains less explored. 
Some researchers~\cite{li2016dexterous,Chen2015adaptive} address stability under uncertainty without employing explicit force control, while others~\cite{li2014learning} learn tactile-reactive grasping but face challenges in generalizing to diverse objects. 
The most relevant work, COP~\cite{takahashi2008COP}, modulates grasp forces based on slip detection and models inter-finger force distribution. 
However, it relies on extensive manual parameter tuning and predefined force allocation rules, limiting its versatility. 
Crucially, most existing works do not explicitly account for rotational slip, which frequently occurs with elongated objects. 
In contrast, our approach leverages a fully actuated hand and a novel SOCP formulation to automatically regulate force distribution and prevent both translational and rotational slip in real time with a unified parameter set across diverse objects.

\

\section{METHOD}

\subsection{Overview}

We first provide a mathematical proof of our key insight: in multi-fingered grasping, preventing translational slip inherently precludes rotational slip (Sec.~\ref{sec:insight}). Next, we introduce our tactile-based Second-Order Cone Programming (SOCP) formulation designed to prevent translational slip (Sec.~\ref{sec:qp}). Finally, we detail the complete tactile-driven control system for compliant and robust grasping (Sec.~\ref{sec:afc}).

\subsection{Geometric Insight for Rotational Stability}
\label{sec:insight}
We now formalize the key geometric insight underlying our approach:
\textit{In a multi-fingered grasp with non-collinear contacts, preventing tangential slip at every contact point is sufficient to preclude global object rotation, without explicitly modeling object-level torque.}

Consider an object grasped at $m$ contact points $\mathbf{o}_i=(x_i,y_i,z_i)^\top$, $i=1,\dots,m$. Without loss of generality, we choose the coordinate frame such that the rotation axis aligns with the $x$-axis. A rotation by an angle $0<\theta<2\pi$ about this axis is then represented by the rotation matrix
\enlargethispage{-\baselineskip}
\[
R(\theta)=
\begin{bmatrix}
1 & 0 & 0\\
0 & \cos\theta & -\sin\theta\\
0 & \sin\theta & \cos\theta
\end{bmatrix}.
\]

\subsubsection{Rigid object}
For a rigid object, after rotation, a point $\mathbf{o}=(x,y,z)^\top$ is mapped to $\mathbf{o}'=R(\theta)\mathbf{o}$, 
while the hand's contact point remains fixed at $\mathbf{o}$, giving the coordinate difference
\[
\Delta \mathbf{d} = \mathbf{o}' - \mathbf{o} = R(\theta)\mathbf{o} - \mathbf{o}
= 
\begin{bmatrix}
0\\
(\cos\theta-1)y - \sin\theta\, z\\
\sin\theta\, y + (\cos\theta-1)z
\end{bmatrix}.
\]
If we prevent tangential slip at this contact, meaning $\Delta \mathbf{d}=\mathbf{0}$, which is equivalent to

\begin{align*}
&\Delta\mathbf{d}=\mathbf{0} \\
\iff& \big((\cos\theta-1)y-\sin\theta z\big)^2
     +\big(\sin\theta y+(\cos\theta-1)z\big)^2=0\\
\iff& \big((\cos\theta-1)^2+\sin^2\theta\big)(y^2+z^2)=0\\
\iff& 4\sin^2\!\tfrac{\theta}{2}\,(y^2+z^2)=0\\
\iff& y=z=0.
\end{align*}
The final equivalence holds because $\theta\not\equiv 0 \pmod{2\pi}$ implies $\sin^2\!\tfrac{\theta}{2}>0$,
so $\Delta \mathbf{d}=\mathbf{0}$ occurs if and only if $y=z=0$, meaning the point lies exactly on the rotation axis.
However, if $\Delta \mathbf{d}=\mathbf{0}$ holds for all contacts, then all contacts must lie on the rotation axis, which contradicts the assumption of non-collinear contacts.
Hence, for a rigid object with non-collinear contacts, preventing tangential slip at every contact precludes any nonzero rotation of the object.

\subsubsection{Deformable object}

For deformable objects, the contact point on the object may deform locally, allowing $\Delta \mathbf{d} = 0$ even when the object rotates, which seem to violate the above insight.
We show that under a commonly used contact modeling assumption—that the deformation at the contact point is dominated by the surface normal—the same conclusion still holds for deformable objects in most practical cases.

Specifically, when the object rotates by $\theta$, we assume that the contact point may deform along the (possibly varying) surface normal direction $\mathbf{n}(\theta)=(n_x(\theta),n_y(\theta),n_z(\theta))^\top$ by an infinitesimal amount $\alpha(\theta)\,d\theta$.

Hence, the new contact point becomes
\[
\mathbf{o'} = R(\theta)\mathbf{o} - \int_{0}^{\theta} \alpha(\tau)\mathbf{n}(\tau)\,d\tau.
\]

Since rotation is continuous, the no-slip condition must hold over an interval rather than at an isolated angle. So  $\exists \theta_0>0,\forall \theta \in [0,\theta_0]$, we require $\Delta \mathbf{d}= \mathbf{0}$, or $\mathbf{o'}= \mathbf{o}$ in other words.

Since both sides are differentiable in $\theta$ and coincide for all $\theta$ in an interval, their derivatives must also coincide. Differentiating with respect to $\theta$ yields
\[
\frac{d}{d\theta}\big(R(\theta)\mathbf{o}\big)
=
\alpha(\theta)\mathbf{n}(\theta),
\]
which can be written explicitly as
\[
\begin{bmatrix}
0\\
-\sin\theta\, y - \cos\theta\, z\\
\cos\theta\, y - \sin\theta\, z
\end{bmatrix}
=
\begin{bmatrix}
\alpha(\theta)\, n_x(\theta)\\
\alpha(\theta)\, n_y(\theta)\\
\alpha(\theta)\, n_z(\theta)
\end{bmatrix}.
\]

Taking the Euclidean norm of both sides yields
\[
\sqrt{y^2+z^2} = \alpha(\theta).
\]

Hence, $\alpha(\theta)=0$ for all $\theta$ if and only if $y=z=0$, i.e., the contact point lies on the rotation axis, which trivially satisfies $\Delta \mathbf{d}=\mathbf{0}$.

We therefore consider the nontrivial case where $y^2+z^2\neq 0$, or in other words, the contact point is not on the rotation axis.
The existence of such a contact point is guaranteed by our assumption of non-collinear contacts.
From $\alpha(\theta)=\sqrt{y^2+z^2}$, it follows that $\alpha(\theta)\neq 0$.
The first component of the equality then gives
\[
n_x(\theta)=0,
\quad
\text{for all } \theta\in[0,\theta_0]
\]
Therefore, the contact normal always lies in the $yz$-plane, normal to the rotation axis.

Eliminating $\alpha(\theta)$ from the remaining two components yields
\[
(-\sin\theta\, y - \cos\theta\, z)\, n_z(\theta)
=
(\cos\theta\, y - \sin\theta\, z)\, n_y(\theta).
\]
This can be rewritten as
$
n_y(\theta)\, y_\theta
+
n_z(\theta)\, z_\theta
=
0$,
where
$(y_\theta, z_\theta)
=
(y\cos\theta - z\sin\theta,\; y\sin\theta + z\cos\theta)$
denotes the coordinates of $(y,z)$ after planar rotation by angle $\theta$. For this equality to hold over an interval, the projected normal $(n_y(\theta), n_z(\theta))$ must remain orthogonal to the continuously rotating vector $(y_\theta, z_\theta)$.

In physical terms, this requires the contact normal to not only stay perpendicular to the rotation axis but also to co-rotate synchronously with the object at every instant. Such an exact, instantaneous synchronization across multiple non-axial contact points corresponds to a highly restrictive geometric configuration. Therefore, in most practical cases, preventing tangential slip at individual contacts effectively precludes global rotational slip, even for deformable objects.

\subsubsection{Summary}
Combining the rigid and deformable analyzes, we conclude that in a multi-fingered grasp with non-collinear contacts, preventing tangential slip at every contact is, in most practical cases, sufficient to preclude global rotational slip of the object.

\begin{figure}[t]
  \centering
  \includegraphics[width=0.8\linewidth]{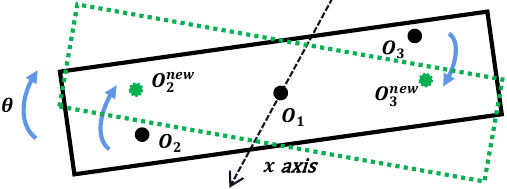} 
  \caption{\textbf{Illustration of our geometric insight: rotational slip inevitably induces translational slip.} 
If the object (black) rotates about the $x$-axis to the dashed pose (green), each $O_i$ moves to $O_i^{\text{new}}$, while $O_1$ stays fixed because it is on the $x$-axis.}

  \label{fig:torque_proof}
\end{figure}

Figure~\ref{fig:torque_proof} provides a geometric illustration of the core mechanism underlying this conclusion. It visualizes how rotational motion about an axis necessarily induces tangential displacement at contact points that do not lie on that axis.

This argument does not extend to parallel grippers, whose typically collinear contacts may align with a rotation axis, allowing rotation without tangential motion.

\subsection{Second-Order Cone Programming with Tactile Feedback}
\label{sec:qp}

As established in the previous section, precluding translational slip is sufficient to prevent global rotational slip. Under the Coulomb friction model, translational slip occurs when contact forces violate the local friction-cone constraints.

This principle directly motivates our SOCP formulation, which computes optimal contact forces that (i) balance the estimated object weight and (ii) strictly reside within the admissible friction cones at all contact points.

\paragraph{Basic formulation}
Consider an object $O$ grasped at $m$ contact points. For each contact $i$, let $\mathbf{n}_i \in \mathbb{R}^3$ denote the inward surface normal, and $\mathbf{d}_i, \mathbf{c}_i \in \mathbb{R}^3$ be orthogonal tangent vectors satisfying $\mathbf{n}_i = \mathbf{d}_i \times \mathbf{c}_i$, all expressed in the world frame.
The Coulomb friction cone $\mathcal{F}_i$ and the local-to-world transformation $\mathbf{J}_i$ are defined as
\begin{align}
\mathcal{F}_i &= \Bigl\{\mathbf{f}_i\in\mathbb{R}^3 ~\Big|~ 
    \gamma_{low}\leq f_{i,1} \leq \gamma_{up},~ 
    f_{i,2}^2+f_{i,3}^2 \leq \tilde{\mu}^2 f_{i,1}^2 \Bigr\}, \label{eq: Fi}\\[3pt]
\mathbf{J}_{i}^\top &=
\begin{bmatrix}
    \mathbf{n}_i & \mathbf{d}_i & \mathbf{c}_i
\end{bmatrix} \in \mathbb{R}^{3\times3}.
\end{align}
Here $\mathbf{f}_i=(f_{i,1},f_{i,2},f_{i,3})^\top$ are the local force components, 
$\tilde{\mu}$ is the estimated friction coefficient (described in the next subsection)
, and $\gamma_{low}=0~N$, $\gamma_{up}=2.5~N$ denote the lower and upper bounds on the normal force.
The small $\gamma_{up}$ is a consequence of the physical limitations of the LEAP hand used in our experiments.

\vspace*{1mm}
To balance the object's weight, the target contact forces 
$\tilde{\mathbf{f}}$ are obtained by solving
\begin{align}
\begin{aligned}
\tilde{\mathbf{f}} 
= &~\underset{(\tilde{\mathbf{f}}_1, \cdots, \tilde{\mathbf{f}}_m)}{\operatorname*{arg\,min}} 
    \big(E_{eq}+\beta_1 E_{smt}+\beta_2 E_{pen}\big) \\
\text{s.t.}\quad & \tilde{\mathbf{f}}_{i} \in \mathcal{F}_i, \quad i=1,\dots,m,
\end{aligned}
\end{align}
where
\begin{equation}
E_{eq}=\Big\|\sum_{i=1}^m \mathbf{J}_{i}^\top \tilde{\mathbf{f}}_{i} - \tilde{\mathbf{G}}\Big\|^2, \,
E_{smt} = \|\tilde{\mathbf{f}} - \tilde{\mathbf{f}}^{\textit{prev}}\|^2, \,
E_{pen} = \|\tilde{\mathbf{f}} \|^2.
\end{equation}
Here, $E_{eq}$ enforces equilibrium by ensuring that the contact forces balance the estimated object's gravitational force $\tilde{\mathbf{G}}\in\mathbb{R}^3$ (described in the next subsection), 
$E_{smt}$ penalizes deviations from the target forces $\tilde{\mathbf{f}}^{\textit{prev}}$ at the previous timestep, and $E_{pen}$ discourages large forces. $\beta_1$ and $\beta_2$ are hyperparameters.

The resulting problem can be efficiently solved using Clarabel~\cite{Clarabel_2024} via \texttt{qpsolvers}~\cite{Caron_qpsolvers_Quadratic_Programming_2025}, 
enabling real-time force computation at around 10 ms.

\paragraph{Incorporating tactile feedback to prevent slip}
Although the solution of the above SOCP formulation satisfies the friction-cone constraints, it differ from the realized contact force, which may exceed the admissible friction-cone bound.
That's because, in implementation, only the normal components are directly regulated via PID tracking (described in the next subsection),
while the tangential forces arise passively from the contact interaction and cannot be actively controlled.

To mitigate this discrepancy, we incorporate tactile feedback to dynamically tighten the lower bound on the normal force in $\mathcal{F}_i$ (Eq.~\ref{eq: Fi}):
\begin{align}
  \gamma_{low} = \frac{\sqrt{(f^{real}_{i,2})^2+(f^{real}_{i,3})^2}}{\tilde{\mu}},
\end{align}
where $(f^{real}_{i,2}, f^{real}_{i,3})$ denote the real-time measured tangential forces at contact $i$.
This adaptive lower bound ensures that the commanded normal force remains sufficient to support the observed tangential load,
thereby compensating for actuation mismatch and maintaining the grasp within the friction cone in practice.

\begin{algorithm}[t]
\caption{Tactile-based Compliant and Robust Grasping}\label{alg: TacDexGrasp}
\begin{algorithmic}[1]  
\Statex \textbf{Data:} $PC$
\Statex \textbf{Param:} model, $n_1$, $n_2$, $\Delta q_{transport}$, $\mu_{init}$, $\mathbf{G}_{init}$
\Statex \textbf{Init:} $\tilde{\mathbf{f}}^{\textit{prev}} \gets \mathbf{0}$
\State $q_{pregrasp}$, $\Delta q_{grasp}$ $\gets$ model($PC$) \Comment{1. Prediction stage}\label{Line: Prediction Stage}

\State \textbf{Move}($q_{pregrasp}$) \Comment{2. Grasping stage}\label{Line: Grasping Stage}
\For{$i = 1$ to $n_1$} \Comment{Squeeze non-contacting fingers}
    \State $mask_{finger} \gets \textbf{GetNonContactFinger}()$
    \State \textbf{MoveDelta}($\Delta q_{grasp} \cdot mask_{finger}$)
\EndFor

\While{True}  \Comment{3. Transport stage}\label{Line: Transport Stage}
    \State \textbf{MoveDelta}($\Delta q_{transport}$)   \Comment{Non-blocking arm control}
    \For{$i = 1$ to $n_2$} \Comment{Adaptive hand control}
        \State $\mathbf{n}$, $\mathbf{f}^{real} \gets \textbf{ReadTactile}()$
        \State $\tilde{\mu}$, $\tilde{\mathbf{G}} \gets \textbf{Update}(\tilde{\mu}, \, \tilde{\mathbf{G}},\,\mathbf{n}, \, \mathbf{f}^{real})$ 
        \State $\tilde{\mathbf{f}} \gets \textbf{SOCP}(\mathbf{n},\, \tilde{\mu},\, \tilde{\mathbf{G}},\, \tilde{\mathbf{f}}^{\textit{prev}})$
        \State \textbf{ContactForceController}($\tilde{\mathbf{f}},\, \mathbf{f}^{real}$)
        \State $\tilde{\mathbf{f}}^{\textit{prev}} \gets \tilde{\mathbf{f}}$
    \EndFor
\EndWhile

\end{algorithmic}
\end{algorithm}

\subsection{Tactile-based Compliant and Robust Dexterous Grasping}
\label{sec:afc}

We now introduce the overall system design for compliant and robust dexterous grasping with tactile feedback, summarized in Algorithm~\ref{alg: TacDexGrasp}, which proceeds through three sequential stages.

\textbf{Prediction Stage (in line~\ref{Line: Prediction Stage})}: This stage predicts grasp poses from the partial observation following~\cite{chen2024bodex, chen2025dexonomy}. Given a single-view segmented object point cloud denoted as $PC$, we use a learned network to predict a pregrasp pose $q_{pregrasp} \in \mathbb{R}^{3+4+J}$,
where the dimensions correspond to 3 for translation, 4 for quaternion rotation, and $J$ for the joint angles. Similarly, the network predicts a delta pose 
$\Delta q_{grasp} \in \mathbb{R}^{3+4+J}$, 
representing the pose adjustment from the pregrasp to the final grasp.

\textbf{Grasping Stage (in line~\ref{Line: Grasping Stage}-6)}: 
In this stage, the hand progressively establishes contact with the object in a compliant manner.
The hand first moves to $q_{pregrasp}$, then gradually closes over $n_1$ timesteps.
At each step, only fingers that are not yet in contact with the object are squeezed (identified via tactile sensing).
This strategy prevents excessive forces at the onset of contact and allows soft adaptation to object geometry.

\begin{figure*}[ht]
  \centering
  \includegraphics[width=1.0\linewidth]{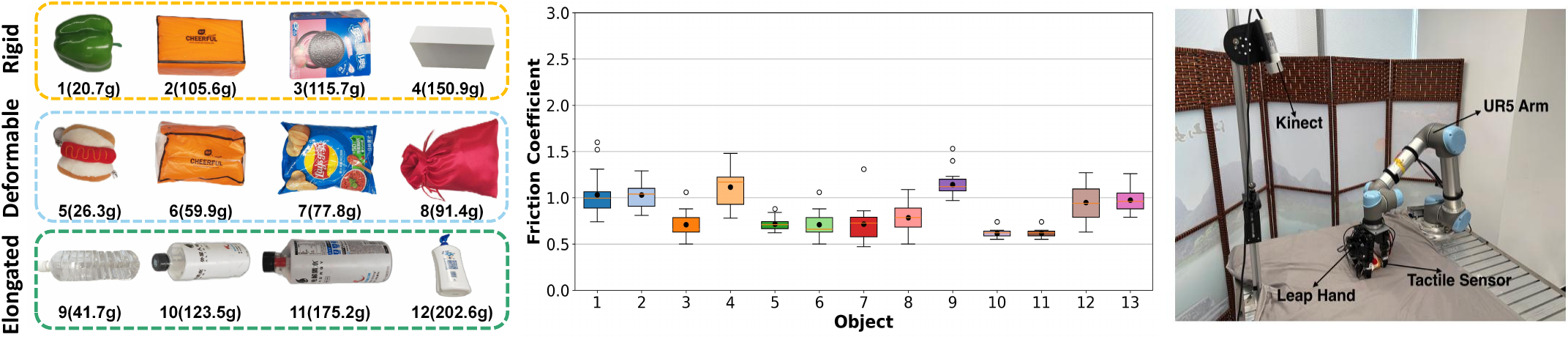} 
  \caption{\textbf{Real World Setup.} (Left) 12 diverse objects. (Middle) Friction coefficients of each object. Note that both mass and friction coefficients are presented only as statistics and are not used in the experiments. (Right) The robotic platform used in our study.}
  \label{fig:leap_setup}
\end{figure*}

\textbf{Transport Stage (in line~\ref{Line: Transport Stage}-16)}:
Once a stable grasp is formed, the hand adjusts its contact forces in closed loop while the arm transports the object along a predefined trajectory, denoted as $\Delta q_{transport}$ at each timestep.
This trajectory typically consists of a gentle lifting phase followed by arbitrary dynamic motions, which can be predefined by users or a high-level planner. 

Our adaptive force control at each timestep is performed as follows:

\begin{itemize}
\item Read tactile data to obtain the contact normals $\mathbf{n}=(\mathbf{n}_1, \cdots, \mathbf{n}_m)$ and the actual contact forces $\mathbf{f}^{real}=(\mathbf{f}^{real}_1, \cdots, \mathbf{f}^{real}_m)$ at each contact point.
\item Update the estimated friction coefficient $\tilde{\mu}$ and gravity $\tilde{\mathbf{G}}$ based on the tactile measurements.
First, the observed friction coefficient at each contact is computed as the ratio between the tangential and normal force magnitudes, and averaged across contacts:
\begin{equation}
    \hat{\mu} 
    = 
    \frac{1}{m} 
    \sum_{i=1}^{m} 
    \frac{\sqrt{(f^{real}_{i,2})^2 + (f^{real}_{i,3})^2}}{f^{real}_{i,1}}.
\end{equation}

Here, $f^{real}_{i,1}$ denotes the normal component of the measured force at contact $i$, while $(f^{real}_{i,2}, f^{real}_{i,3})$ are its tangential components in the local contact frame.

Second, the object gravitational force is estimated from the measured contact wrenches:
\begin{equation}
    \hat{\mathbf{G}} 
    =
    \frac{g}{g+a}
    \sum_{i=1}^{m} 
    \mathbf{J}_{i}^\top \mathbf{f}^{real}_i,
\end{equation}
where $\mathbf{J}_{i}^\top$ maps the local contact force $\mathbf{f}^{real}_i$ to the world frame, 
$g$ is the gravitational acceleration ($9.8~\text{N/kg}$), and $a$ is the object acceleration estimated from the hand kinematics. 
The factor $\tfrac{g}{g+a}$ compensates for inertial effects so that $\hat{\mathbf{G}}$ approximates the true gravity rather than the total dynamic wrench.

To improve robustness against sensor noise and transient disturbances, we apply a maximum sliding-window filter to $\hat{\mu}$ and an average sliding-window filter to $\hat{\mathbf{G}}$, yielding the final estimates $\tilde{\mu}$ and $\tilde{\mathbf{G}}$.

Additionally, we set initial estimates $\tilde{\mu}_{init}=0.4$ and $\tilde{\mathbf{G}}_{init}=20g$,
which are reasonable lower bounds for the friction coefficient and object weight, respectively, 
and allow the system to start with a small initial normal force to gather tactile feedback for subsequent estimation updates.

\item Using our tactile-based SOCP formulation introduced in Section~\ref{sec:qp} to compute the target contact forces.
    \item Using a joint position-based PID controller to track the target forces solved by SOCP:
\begin{equation}
\begin{aligned}
q_{control} &= q_{current} + k_p \, \tau + k_i \int \tau \, dt + k_d \, \dot{\tau}
\end{aligned}
\end{equation}
where $\tau=\mathbf{J}^\top(q_{current}) (\tilde{\mathbf{f}} - \mathbf{f}^{real})$ is the error torque and $\mathbf{J}^\top(q_{current}) \in \mathbb{R}^{J \times 3m}$ is the hand Jacobian mapping the stacked contact forces at \(m\) contact points to the joint torques.Here, $J$ denotes the number of joints in the robot hand.

\end{itemize}

\section{EXPERIMENT}
\subsection{Evaluation Metrics}

The following metrics are adopted for evaluation:
\begin{itemize}
    \item \textbf{Success Rate}: The ratio of successful grasps to the total number of trials. A trial is considered successful if the object is lifted, transported for 5 seconds without visible slip or drop.
    \item \textbf{Relative Max Force ($F_{max}/G$)}: The peak contact force applied to the object $F_{max}$, normalized by its weight $G$, enabling comparison across objects of varying masses.
    \item \textbf{Relative Gravity Estimation ($\tilde{G}/G$)}: The ratio of the estimated gravity $\tilde{G}$ to the ground-truth gravity $G$. Values close to 1 indicate accurate estimation.
    \item \textbf{Relative Friction Coefficient Estimation ($\tilde{\mu}/\mu$)}: The ratio of the estimated friction coefficient $\tilde{\mu}$ to the ground-truth $\mu$. Values close to 1 indicate accurate estimation.
\end{itemize}

\subsection{Experiment Setup}

\textbf{Hardware Platform:}
Our hardware setup consists of a 16-DoF Leap hand mounted on a 6-DoF UR5e robotic arm, complemented by an Azure Kinect sensor for RGB-D perception.
Additionally, four Tac3D visual-tactile sensors~\cite{zhang2022tac3d} are mounted on each fingertip, operating at a sampling rate of 30 Hz.
Each Tac3D sensor generates a $10 \times 10$ taxel array, where each taxel captures a 3D force vector. We derive the resultant contact force for each fingertip by aggregating these spatial force distributions into a single 3D vector. This representation is sufficient for our control law as contact is localized within the fingertip region.

\textbf{System Frequency:}
While our SOCP-based PID controller is capable of running at 100 Hz, the effective closed-loop update rate is 30 Hz, constrained by the tactile sensing pipeline.

\textbf{Object Set:}
As illustrated in Figure~\ref{fig:leap_setup}, 
we evaluate 12 objects with varying material deformability, friction coefficients, and mass distributions. 
The object masses range from 20g to 200g, and friction coefficients range from 0.4 to 1.2, 
measured by gradually opening from a parallel grasp configuration and recording the force at which the object slips.
All estimated gravity and friction coefficients in the figure are presented as statistics for analysis, but are not used in the experiments.

For baselines and ablation studies, the 12 objects are grouped into three categories:
\begin{itemize}
    \item \textbf{Rigid objects:} Non-deformable objects with regular shapes (Apple, Tissue, Oreo Box, White Box).
    \item \textbf{Deformable objects:} Objects with deformable materials (Soft Toy, Half Tissue, Chips, Candy Pack).
    \item \textbf{Elongated objects:} Long objects that are prone to rotational slip (Juice Pack,  Bottle, Heavy Bottle, Shampoo).
\end{itemize}

\textbf{Experimental Protocol:}
Each object is tested in five trials, leading to 60 trials per method (including each baseline and ablation setting), yielding a total of 300 trials in the real world.
In each trial, we first segment the object in the RGB image using Segment Anything~\cite{ravi2024sam} to obtain a segmented point cloud, 
and then execute Algorithm~\ref{alg: TacDexGrasp} to perform the grasp.

\subsection{Comparison with Baselines}
\begin{table}[t]
\centering
\setlength{\tabcolsep}{5pt}
\begin{tabular}{@{}lcc|cc|c@{}}
\toprule
 & \multicolumn{2}{c|}{\textbf{Baselines}} & \multicolumn{2}{c|}{\textbf{Ablations}} & \multirow{2}{*}{\textbf{Ours}} \\
\cmidrule(lr){2-3} \cmidrule(lr){4-5}
\textbf{Object} & COP & BODex & w/o PID & w/o SOCP & \\
\midrule
\textbf{Rigid} & 2.01/70\% & 2.11/\textbf{100\%}& \textbf{1.57}/90\% & 1.58/60\% & 1.58/95\% \\
\textbf{Deformable}   & 2.38/55\% & 3.62/70\% & 1.79/60\% & 2.20/35\% & \textbf{1.24}/\textbf{75\%} \\
\textbf{Elongated} & 3.29/25\% & 3.56/45\% & 3.14/40\% & 2.83/20\% & \textbf{1.96}/\textbf{80\%} \\
\midrule
\textbf{Total}  & 2.56/50\% & 3.10/72\% & 2.17/63\% & 2.20/38\% & \textbf{1.59}/\textbf{83\%} \\
\bottomrule
\end{tabular}
\caption{\textbf{Baseline comparison and ablation study.} Each entry reports the relative max force $F_{max}/G$ (lower is better) and success rate (higher is better). Our method achieves the best overall trade-off, particularly on deformable and elongated objects. Removing either the PID or SOCP module leads to notable degradation.}
\label{tab:baseline results}
\end{table}
\begin{figure*}[!htbp]
  \centering
  \includegraphics[width=1.0\linewidth]{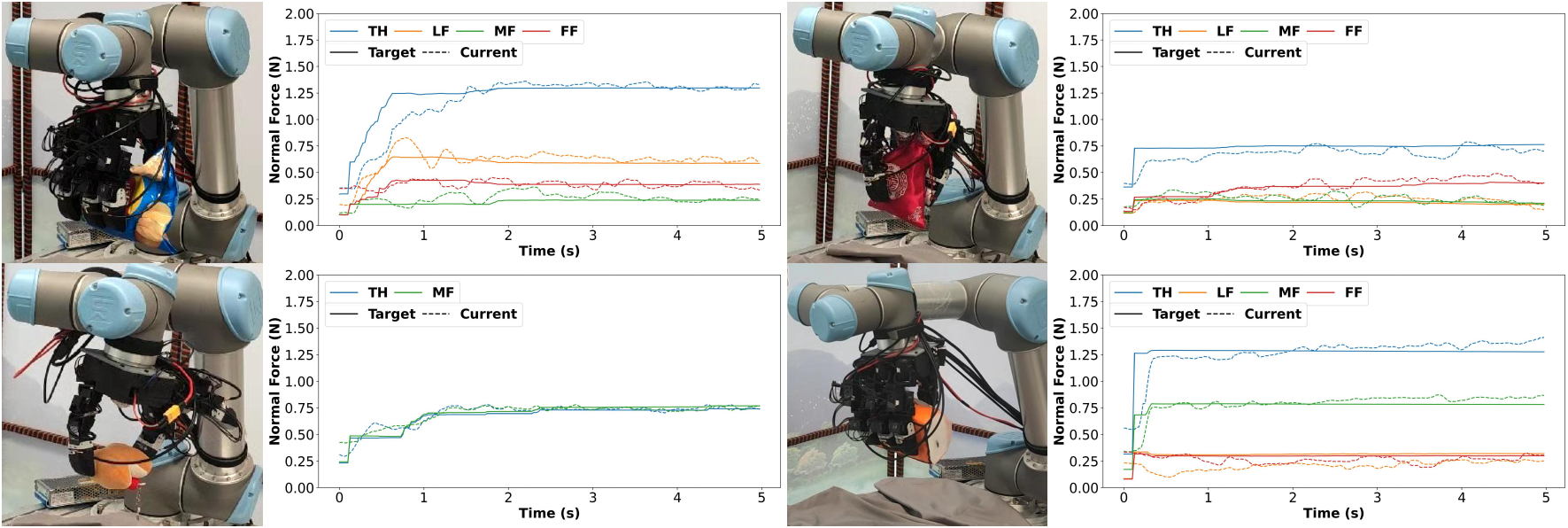}
  \caption{\textbf{Real-World Experiments.} 
  For diverse objects with varying masses, deformability, and friction, our system rapidly adapts and achieves stable grasps.
  The measured contact forces align closely with the SOCP-derived target forces. 
  Legends denote fingertip labels: TH (thumb), LF (little finger), MF (middle finger), and FF (forefinger).}
  \label{fig:pid_figure}
  \vspace{-2mm}
\end{figure*}

The following baseline methods are considered:

\begin{itemize}
    \item \textbf{BODex}~\cite{chen2024bodex}: Computes a squeeze grasp pose via $q_{\text{squeeze}} = q_{\text{pregrasp}} + \eta \Delta q_{\text{grasp}}$ 
    and directly executes this pose as the control target. We empirically set $\eta = 3$ for the best performance.
    \item \textbf{COP}~\cite{takahashi2008COP}: Squeezes when contact point slip and relaxes once the contact point stabilizes.
    COP requires a distinct parameter set for each object and grasp pose; for fairness, we used a unified set that yielded the best overall performance across all objects.
\end{itemize}

Table~\ref{tab:baseline results} shows that our method consistently outperforms all baselines in both success rate and relative maximum force.

BODex lacks tactile-based regulation for force magnitude or inter-finger distribution, making it unable to adapt to variations in object stiffness and geometry. Consequently, it often applies excessive force to rigid objects while failing to provide balanced stabilizing forces for deformable or elongated ones, resulting in lower success rates and higher peak contact forces.

COP achieves better force magnitudes than BODex via slip-triggered adjustments. However, its heuristic adaptation lacks an explicit mechanism for coordinated inter-finger force distribution. This leads to suboptimal force allocation, particularly for deformable objects requiring continuous redistribution under shape changes. Furthermore, since COP does not explicitly model or constrain rotational slip, it struggles with elongated objects that demand precise torque balancing.
\subsection{Ablation Study}

For the ablation study, we disable key modules one at a time:
\begin{itemize}
    \item \textbf{w/o PID:} The PID controller is removed, and the desired contact forces are applied directly.
    \item \textbf{w/o SOCP:} The SOCP formulation is omitted; the thumb applies a force $\eta \tilde{G}/(2\tilde{\mu})$, 
    while the remaining fingers equally share the residual force $F_{other}=\frac{\eta \tilde{G}}{2(n-1)\tilde{\mu}}$. 
    We empirically set $\eta=2$ for best performance.
\end{itemize}

Results in Table~\ref{tab:baseline results} demonstrate that removing either module substantially degrades performance—success rates drop, and peak contact forces increase consistently across all object types.

Without the PID controller, the system lacks closed-loop tracking of the desired contact forces. 
Execution errors accumulate over time due to actuation delays and model mismatch, causing deviations from the intended force profile. 
These deviations are particularly detrimental for elongated objects, where even small tracking errors can generate unbalanced torques and induce rotational slip.

Without the SOCP module, contact forces are assigned according to a fixed heuristic distribution rather than being optimized under friction-cone and equilibrium constraints. 
So the hand cannot dynamically coordinate inter-finger forces to maintain rotational stability or adapt to object-dependent variations. 
This leads to imbalanced force allocation, increased peak contact forces, and reduced grasp robustness, yielding performance comparable to heuristic baselines such as COP.
\subsection{Detailed Result and Failure Analysis}

\begin{table}[t]
\centering
\setlength{\tabcolsep}{5pt}
\begin{tabular}{@{}clcccc@{}}
\toprule
\textbf{Category} & \textbf{Object} & $\tilde{G}/G$ & $\tilde{\mu}/\mu$ & $F_{max}/G$ & \textbf{Success}($\#$/5) \\
\midrule
\multirow{4}{*}{\textbf{Rigid}} 
 & 1-Apple & 1.20 & 0.88 & 2.01 & 4/5 \\
 & 2-Tissue & 1.10 & 0.81 & 1.83 & 5/5 \\
 & 3-Oreo Box & 1.06 & 0.86 & 1.38 & 5/5 \\
 & 4-White Box & 0.98 & 0.60 & 1.11 & 5/5 \\
\midrule
\multirow{4}{*}{\textbf{Deformable}} 
 & 5-Soft Toy & 0.82 & 1.11 & 1.03 & 5/5 \\
 & 6-Half Tissue & 1.11 & 0.90 & 1.35 & 3/5 \\
 & 7-Chips & 1.03 & 1.44 & 1.16 & 4/5 \\
 & 8-Candy Pack & 0.89 & 0.67 & 1.41 & 3/5 \\
\midrule
\multirow{4}{*}{\textbf{Elongated}} 
 & 9-Juice Pack & 1.09 & 0.96 & 2.99 & 4/5 \\
 & 10-Bottle & 1.12 & 1.21 & 1.90 & 5/5 \\
 & 11-Heavy Bottle & 1.01 & 1.14 & 1.55 & 4/5 \\
 & 12-Shampoo & 1.22 & 0.90 & 1.39 & 3/5 \\
\bottomrule
\end{tabular}
\caption{\textbf{Detailed quantitative results across object categories.} 
Our method achieves high success rates while adaptively maintaining low contact forces across all objects.}

\label{tab:results}
\end{table}

Table~\ref{tab:results} reports per-object results. 
Overall, TacDexGrasp achieves an average success rate of 83\% across all objects, while keeping maximum contact forces below $2G$ in most cases,
indicating that stability is achieved through coordinated force distribution rather than brute-force compression.
The estimated gravity ratios $\tilde{G}/G$ remain close to 1 across objects,
demonstrating reliable online weight estimation.
Similarly, friction estimation stays within a reasonable range,
allowing the SOCP solver to maintain contact forces inside friction cones. 

Figure~\ref{fig:pid_figure} visualizes representative trials on four objects, showing the target contact forces derived from the SOCP formulation alongside the actual forces recorded by tactile sensors.  
The results highlight how our method starts with low contact force, gradually increases it based on tactile feedback, and converges to stable target values.  

Failure cases primarily arise for deformable and elongated objects. 
For deformable objects (e.g., Half Tissue and Candy Pack), 
significant shape deformation can shift contact regions beyond the sensing coverage of the tactile array, 
leading to partial or complete loss of feedback and subsequent slip. 
For elongated objects (e.g., Shampoo), failures are associated with larger mass and moment arms between contact locations and the center of mass. 
Such conditions demand rapid and substantial force adjustments to maintain rotational equilibrium. 
In some cases, the required response exceeds the system limits (sensing bandwidth constrained by the 30\,Hz tactile update rate and actuation capacity bounded by the maximum output force of the Leap hand), resulting in instability and slip.
\subsection{Robustness Test}
\begin{figure}[t]
  \centering
  \includegraphics[width=1.0\linewidth]{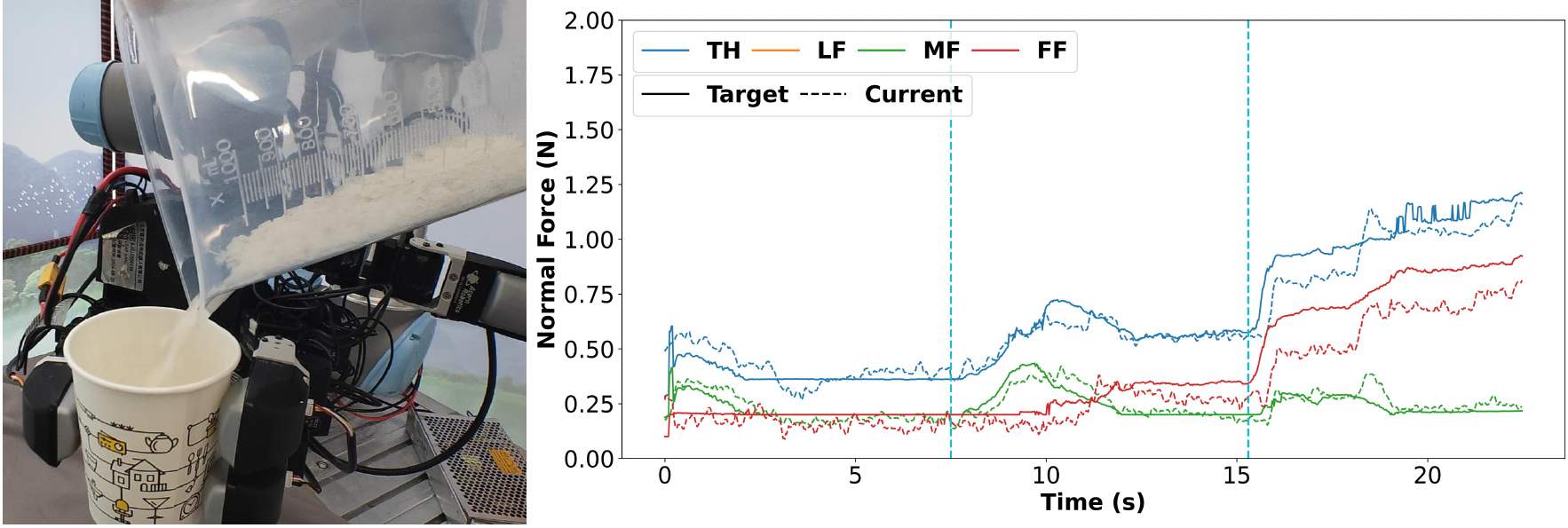}
  \caption{\textbf{Mass Adaptation Test.} Our system quickly responds to two sudden mass increases by increasing the target normal forces accordingly.}
  \label{fig:mass_adaption}
\end{figure}

\begin{figure}[t]
  \centering
  \includegraphics[width=1.0\linewidth]{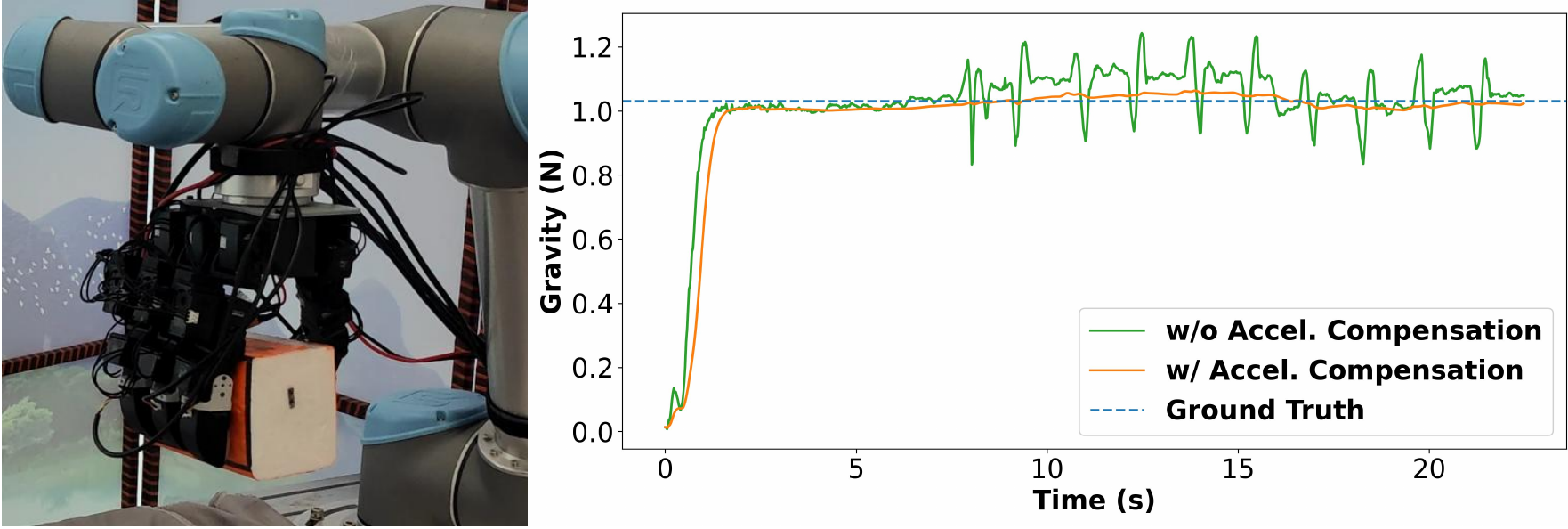}
  \caption{\textbf{Rapid Shaking Test.}
Our system accurately estimates object weight during vigorous shaking by compensating for acceleration-induced disturbances.}
  \label{fig:shake}
\end{figure}

\begin{figure}[t]
  \centering
  \includegraphics[width=0.9\linewidth]{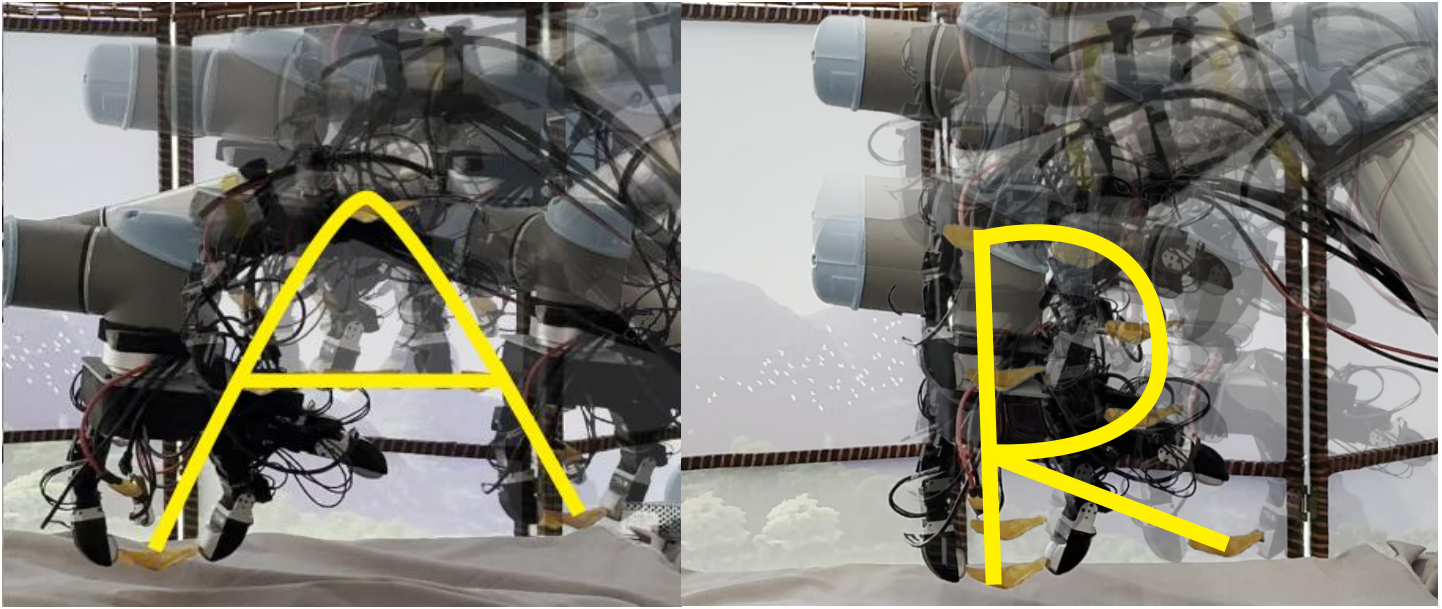}
  \caption{\textbf{Irregular Motion Test.} The system holds a fragile crisp through irregular motions without causing damage.}
  \label{fig:irregular}
\end{figure}
Beyond average metrics, we further examine the robustness of TacDexGrasp under challenging real-world scenarios. 
We design three case studies covering mass variation, rapid shaking, and irregular arm motion.

\textbf{Mass Adaptation Test.} 
To simulate mass changes during manipulation (e.g., filling a container), we grasp a paper cup and pour rice into it twice at distinct moments.
As shown in Fig.~\ref{fig:mass_adaption}, each sudden weight increase is quickly detected through tactile feedback, and the controller promptly strengthens normal forces to maintain stability without deforming the cup.

\textbf{Rapid Shaking Test.} 
We further evaluate the system under strong inertial disturbances by vigorously shaking a grasped bottle.
As shown in Fig.~\ref{fig:shake}, TacDexGrasp maintains stable gravity estimation despite large accelerations, effectively compensating for inertial effects and preventing slip throughout the motion.

\textbf{Irregular Motion Test.} 
We further evaluate grasp robustness by having the arm trace the letters “A” and “R” in the air while holding a fragile crisp, as shown in Fig.~\ref{fig:irregular}.
Despite abrupt and irregular motions, the grasp remains stable without damaging the crisp, demonstrating the controller's compliance and robustness.

Overall, TacDexGrasp demonstrates strong robustness to mass variation, rapid perturbation, maintaining both safety and stability in dynamic settings.

\section{LIMITATION AND FUTURE WORK}
First, the range of object weights we can manipulate is restricted by the limited force output of the Leap hand.  
Future work could investigate stronger dexterous hands to expand the range of manipulable object masses.  
Second, the diversity of grasp poses is constrained by the tactile sensing area, which is currently limited to the fingertips.  
Equipping tactile sensors on the palm or finger sides could enable richer contact interactions and support more versatile grasp strategies, as suggested in Dexonomy~\cite{chen2025dexonomy}.  
Finally, to better assess the generalization capability of our method, more diverse objects and more complex manipulation tasks (e.g., tool use) are expected to be validated in the future work.   

\section{CONCLUSION}
We present \textbf{TacDexGrasp}, a tactile-feedback-driven SOCP controller for compliant and robust dexterous grasping.
By constraining tangential-to-normal force ratios, it prevents both translational and rotational slip without explicitly modeling gravitational torque.
Real world experiments on diverse objects demonstrate higher success rates and lower peak forces than previous baselines. We also show the robustness of our method to mass changes and dynamic disturbances.
\addtolength{\textheight}{-12cm}   








References are important to the reader; therefore, each citation must be complete and correct. If at all possible, references should be commonly available publications.

\bibliographystyle{IEEEtran}
\bibliography{IEEEabrv}

\end{document}